\documentclass[letterpaper, 10 pt, conference]{ieeeconf}

\IEEEoverridecommandlockouts
\usepackage{cite}
\usepackage{times}
\usepackage{mwe}
\usepackage{multicol}
\let\labelindent\relax % ieeeconf.cls already defines \labelindent; avoid clash with enumitem
\usepackage{enumitem}
\usepackage{algorithm}
\usepackage{algpseudocode}
\usepackage{graphicx}
\usepackage{amsmath}
\usepackage{amssymb}
\usepackage{tikz}
\usepackage{relsize}
\usepackage{booktabs}
\usepackage[font={small}]{caption}
\usepackage[table]{xcolor}
\usepackage{xspace}
\usepackage{pifont}
\usepackage{colortbl}
\usepackage[bookmarks=true,colorlinks=true]{hyperref}

\usepackage[table]{xcolor}
\usepackage{booktabs}
\usepackage{multirow}
\usepackage{graphicx}

\usepackage{xfp}
\definecolor{srlow}{HTML}{FFFCF0}
\definecolor{srhigh}{HTML}{FFD400}
\newcommand{\srfrac}[2]{\cellcolor{srhigh!\fpeval{round(100*(#1/#2)^3)}!srlow}#1/#2}
\newcommand{\sr}[1]{\srfrac{\fpeval{#1/10}}{10}}

\newcommand{\method}{Steertuning}
\newcommand{\website}{\url{https://steertuning.github.io}}

\definecolor{blue}{HTML}{0173B2}
\definecolor{green}{HTML}{06A66C}
\definecolor{red}{HTML}{CF3C33}
\definecolor{magenta}{HTML}{D748A2}

\hypersetup{
    colorlinks=true,
    linkcolor=blue,
    filecolor=magenta,
    citecolor=blue,
    urlcolor=teal,
}

\makeatletter
\def\bstctlcite{\@ifnextchar[{\@bstctlcite}{\@bstctlcite[@auxout]}}
\def\@bstctlcite[#1]#2{\@bsphack
  \@for\@citeb:=#2\do{%
    \edef\@citeb{\expandafter\@firstofone\@citeb}%
    \if@filesw\immediate\write\csname #1\endcsname{\string\citation{\@citeb}}\fi}%
  \@esphack}
\makeatother

\title{
    \LARGE \bf
    Steertuning Policies from a Few Human Videos
}

\author{
Billy Yan \quad Isaac Tu$^{*}$ \qquad Kelly Lee$^{*}$ \qquad Blake Chang \qquad Kenny Ma \qquad Kanad Patel \\ 
Dhawal Kabra \qquad Neer Patel \qquad Kevin Wu \qquad Lerrel Pinto \qquad Irmak Guzey
\\ \\ New York University
\\ \\ {\tt \href{https://steertuning.github.io}{steertuning.github.io}}\\
\thanks{Correspondence to \texttt{irmakguzey@nyu.edu} and \texttt{yy4435@nyu.edu}. ${*}$ denotes equal contribution.}
}

\makeatletter

\let\@oldmaketitle\@maketitle

\renewcommand{\@maketitle}{%
    \@oldmaketitle
    \centering
    \includegraphics[width=\linewidth]{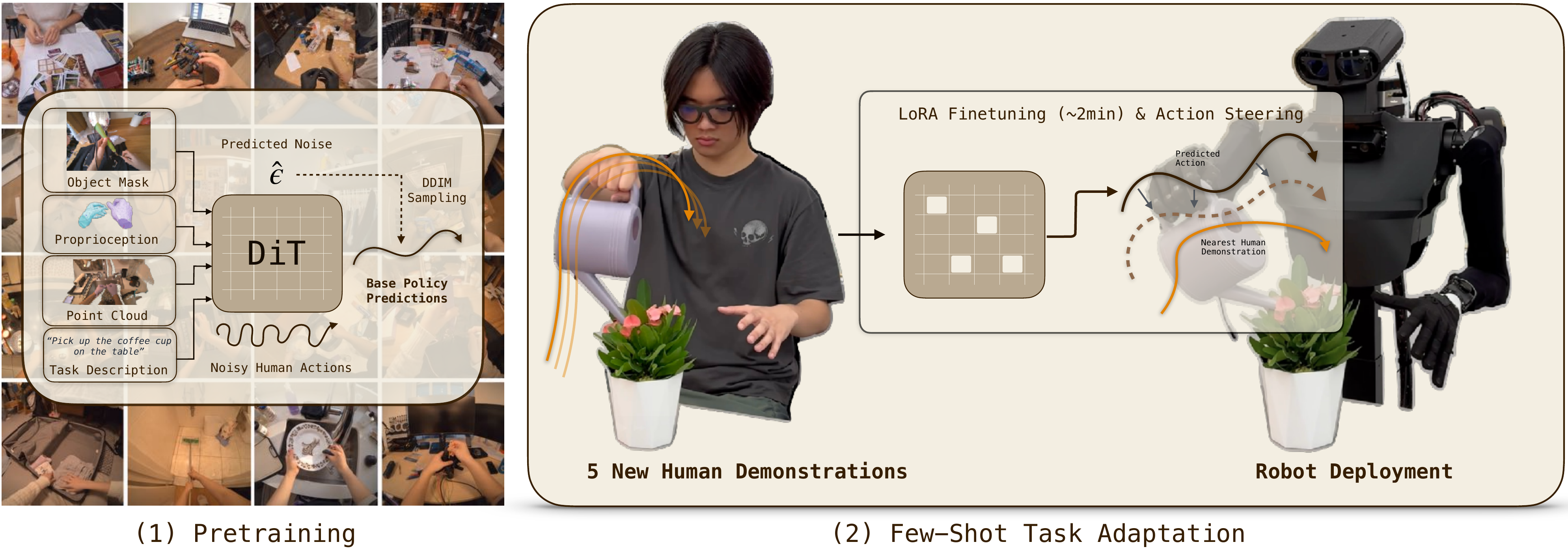}
    \captionof{figure}{
        \method{} is an end-to-end system for few-shot task generalization in completely new environments. It consists of two parts: (1) pretraining a 400M-parameter diffusion policy on our human dataset consisting of 100 hours of densely annotated random interactions, and (2) 2-minute finetuning and diffusion steering, used together on 5 new task demonstrations collected with Aria Gen 2 glasses. By carefully combining the two, our mobile manipulator completes 12 new tasks in 12 new environments in under 5 minutes with a $71\%$ average success rate.
    }
    \label{fig:intro}
    \vspace{-5pt}
}

\makeatother

\begin{document}
\bstctlcite{IEEEexample:BSTcontrol}

% Notes
\newcommand{\xxnote}[3]{}

\ifx\hidenotes\undefined
    \renewcommand{\xxnote}[3]{\color{#2}{#1: #3}}
\fi

\newcommand{\IG}[1]{%
    {\xxnote{IG}{magenta}{#1}}%
}

% Title + teaser
\maketitle
% Abstract
\begin{abstract}
    Current recipes for robot learning require massive amounts of robot demonstrations to obtain robust and adaptive policies. In this work, we introduce \method{}, a system that enables robots to learn new tasks in new environments from 5 human demonstrations in under 5 minutes.

\method{} consists of two parts. (1) We pretrain a 400M-parameter diffusion policy on our egocentric human dataset, which consists of 100 hours of densely annotated random interactions. This pretraining makes \method{} robust to background differences and enables spatial and object generalization when deployed on new tasks. (2) Given 5 demonstrations of a new task, we finetune our model for only 2 minutes (2000 training updates) and steer the policy's actions during reverse sampling using the demonstration whose observation is closest to that of the deployment environment. Together with a carefully designed data processing pipeline for embodiment transfer, this enables \method{} to achieve a $71\%$ success rate across 12 tasks in 12 new environments, ranging from lab settings to kitchens in new homes. The entire process, from data collection for a new task to robot policy deployment, takes less than 5 minutes. Our full dataset, model weights, codebase, and robot videos will be made available at \website{}.
\end{abstract}

\IEEEpeerreviewmaketitle

% Main paper
\section{Introduction}

Humans learn new tasks every day: we adapt and improve through new interactions and by watching others. Achieving similar flexibility in robots requires systems capable of the same kind of adaptation. Recent work on robot learning from human videos has enabled many ways of using scalable human demonstrations for downstream robotic tasks. Some combine robot and human data to enable quick adaptation~\cite{dexwild, in-n-on}, where models trained on large-scale human data adapt to new tasks from a few robot demonstrations. Others learn few-shot robot policies from human demonstrations alone~\cite{egozero,aina,point-policy}. However, learning representations that are both scalable and robust, while adapting quickly from human data only, remains unexplored. In this work, we introduce \method{}, an end-to-end mobile dexterous manipulation system trained and finetuned only on egocentric human data, which adapts to completely new environments and homes within 5 minutes.

Initial efforts toward general robot policies focused on behavior cloning (BC), where a robot is teleoperated~\cite{openteach, holodex, bunny-pro} to complete tasks under different configurations and the recorded data is used to train a policy that clones these behaviors. However, this requires hundreds, if not thousands, of demonstrations per task. To reduce the data requirement per task, past work showed that pretraining policies on very large datasets~\cite{pi0, octo, openx} and then \textit{finetuning} a small subset of the parameters~\cite{lora} on task-specific demonstrations can boost performance. Yet, for large enough models, this process can still take days. Alternatively, with denoising diffusion probabilistic models (DDPMs), a desired-yet-incomplete distribution can be used to \textit{steer} the predicted distribution toward it~\cite{repaint}, i.e., inpainting with a provided mask. Similar approaches have succeeded in robotics as well~\cite{demo-diffusion}: given a large pretrained diffusion model and a few task demonstrations, steering enables task adaptation while keeping the deployed policy safe and robust. \method{} carefully combines finetuning and action steering, so that adaptation takes only 2 minutes and reaches an $83\%$ success rate across four of our tasks, compared to $23\%$ with steering only and $0\%$ with finetuning only (Section~\ref{sec:experiments:steertuning}).

Meanwhile, given the challenges of teleoperation, human data is receiving increasing attention in the robot learning community. Initial efforts collected human demonstrations in a calibrated robot environment with multiple cameras, enabling accurate 3D annotations such as hand poses and object points~\cite{point-policy, hudor, dexcap, demo-diffusion}. While these works show the importance of 3D annotations for human-to-robot retargeting, a calibrated environment is too strong an assumption for general-purpose home robots. Others instead leveraged internet-scale egocentric human videos~\cite{vrb, track2act, zeromimic}, showing how scalable data can improve robots' understanding of interactions; however, the lack of annotations such as 3D hand poses limited their deployability. More recent works~\cite{egozero, egomimic, aina, hug} show that smart glasses (e.g., Aria Gen 1/2) are a sweet spot in this scalability--annotation trade-off, enabling deployment of robot policies trained purely on human data.

Building on this line of work, \method{} combines the scalable, richly annotated data from smart glasses with the test-time adaptation capabilities of finetuning and action steering. As illustrated in Fig.~\ref{fig:intro}, \method{} consists of two parts. The first is pretraining a 400M-parameter diffusion transformer on our random-interaction dataset; 100 hours of egocentric Aria Gen 2 recordings of humans performing everyday tasks. The transformer takes environment observations and a task description as input and predicts the noise $\epsilon$ added to the human actions. The second is 2-minute finetuning and action steering, used together on 5 new egocentric demonstrations, which enables adaptation to a new task in a new environment in under 5 minutes total. Our contributions are:
\begin{enumerate}
    \item \method{} is a new recipe that combines finetuning and action steering to enable quick yet robust task adaptation for diffusion policies. We ablate the design decisions behind it and show a $71\%$ success rate across 12 tasks in 12 new environments (Section~\ref{sec:method:adaptation}).
    \item To our knowledge, \method{} is the first end-to-end system trained purely on human data that robustly adapts to new tasks in completely new environments (Section~\ref{sec:method:overview}).
    \item We integrate \method{} with an open-knowledge navigation system~\cite{ok-robot} and a VLM orchestrator~\cite{gemini} to deploy robot policies from a single language command, showing that \method{} generalizes even when navigation and docking are handled by our mobile manipulator.
\end{enumerate}

Our dataset, model weights, robot videos, and software stack will be made available at \website{}.

\section{Related Work}

\subsection{Fine-Tuning and Steering Pretrained Policies}

\textbf{Fine-tuning.} The dominant recipe for few-shot task learning is to pretrain a large policy on open-source robot datasets~\cite{holodex,openteach,openx,openvla,rdt} and post-train it on a small set of task-specific demonstrations. Since full fine-tuning is too flexible for a handful of demonstrations, parameter-efficient methods such as low-rank adapters are used instead~\cite{lora,tail,openvla-oft}, and others fine-tune diffusion policies with RL~\cite{dppo,dsrl}. In all cases, adaptation requires new robot data and, for large models, hours to days of training.

\textbf{Steering.} A complementary line of work adapts a pretrained diffusion model at inference time, without changing its weights, by modifying the reverse sampling process. This started in image generation with guidance~\cite{cfg} and inpainting~\cite{repaint}, and was later adopted by diffusion planners~\cite{diffuser,safediffuser,mpd} and diffusion policies~\cite{diffusion-policy}, where sampling is steered with a value function, a dynamics model, or human corrections~\cite{vgps,dynaguide,itps}. Closest to ours is DemoDiffusion~\cite{demo-diffusion}, which steers a pretrained diffusion policy with a single retargeted human demonstration. Steering alone, however, can only bend the policy toward the demonstration; it cannot teach behaviors the base policy does not know. \method{} instead combines a 2-minute LoRA fine-tuning with demonstration-based steering, which we show to be both faster and more robust than either alone (Section~\ref{sec:experiments:steertuning}).

\subsection{Policy Learning from Human Data}

Learning robot policies from human data can be split by the data source used. Initial work used in-domain data, where humans demonstrate tasks in the same calibrated environment as a table-top manipulator~\cite{aina,demo-diffusion,hudor,dexcap}. While effective, these approaches are not scalable, since robots are no longer stationary. On the other hand, internet-scale in-the-wild videos are scalable but lack annotations needed for robot deployment, such as 3D hand pose or depth. Past work that used internet videos therefore either extracted higher-level interaction information such as affordances~\cite{vrb,hap,swim} or object tracks~\cite{track2act,zeromimic,atm,robotap}, or first retargeted human demonstrations to robots~\cite{dexmv,videodex,mimicplay,phantom,motion-tracks} and then trained policies on the retargeted trajectories. The first option lacks the low-level annotations needed for dexterous interactions, and the second requires extensive preprocessing for every video with a high failure rate. More recently, human videos have also been used to pretrain policies that are then fine-tuned on robot data~\cite{lapa,egovla,humanoid-human,dexwild,in-n-on}, but this still requires robot demonstrations for every new task.

Smart glasses provide a sweet spot between these two branches. While enabling in-the-wild, scalable data collection, they provide useful 3D annotations such as head and hand pose and stereo depth. Past work showed promising results for learning robot policies from data collected with Aria glasses~\cite{egomimic, egozero, aina}. But these works either only worked for parallel grippers, relied on point tracks of the objects, or lacked the object generality needed for general robots. \method{} introduces a new system that (a) works end-to-end in new environments and homes, showcasing extensive visual generalization, (b) can adapt to new tasks with under 5 minutes, and (c) uses data collected purely from humans wearing Aria Gen 2 glasses.
\section{Method}
\label{sec:method}

\subsection{Overview}
\label{sec:method:overview}

\method{} is an end-to-end mobile dexterous manipulation system that completes tasks from language commands and adapts to new tasks in under 5 minutes of processing time. The system consists of three parts: (a) collecting and processing egocentric human data for scalable and robust robot learning, (b) learning robot policies from this data that robustly adapt to new environments and tasks, and (c) closing the human-to-robot embodiment gap and combining the policies with open-knowledge navigation~\cite{ok-robot} so that new tasks can be completed from a single language command. We describe each part below.

\begin{figure*}[t]
  \centering
  \includegraphics[width=\linewidth]{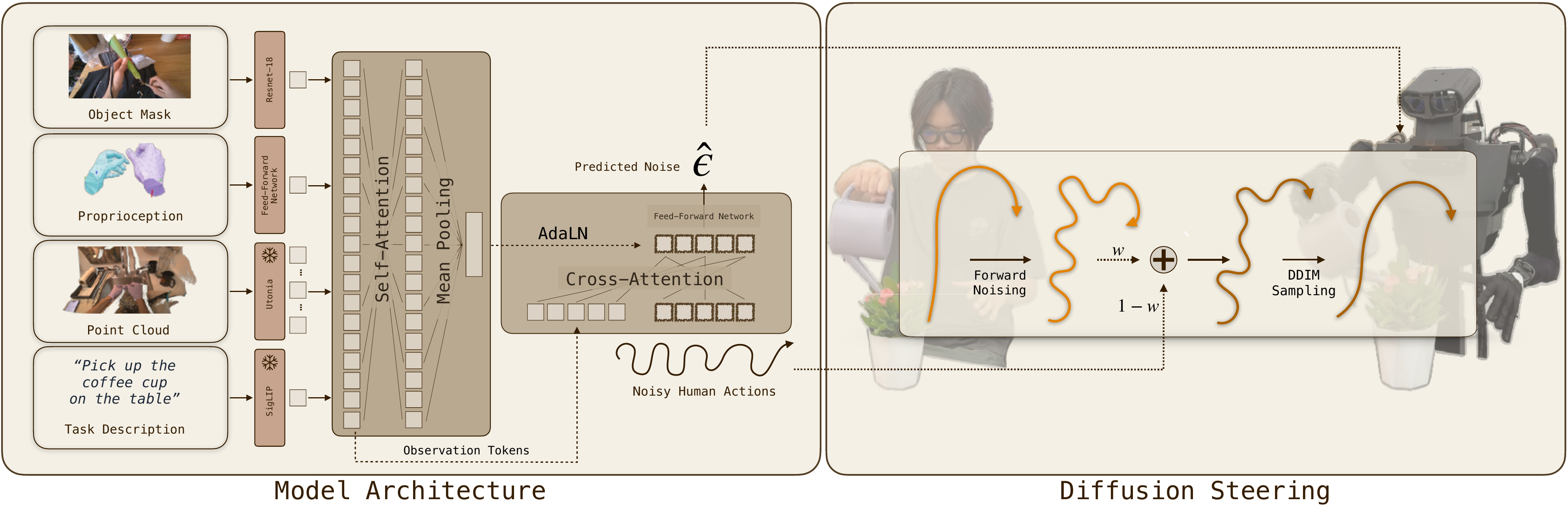}
  \caption{\textbf{Policy.} Our architecture and action steering mechanism. The policy takes four modalities as input: a point cloud, the manipulated-object mask, the fingertip locations of both hands (10 3D points) in the anchor frame, and a text description of the task. Each modality is passed through its own encoder and the resulting tokens go through a self-attention transformer, after which we mean-pool them into a \textit{context latent} that conditions the diffusion transformer via AdaLayerNorm~\cite{adaln}. The observation tokens are also fed to the diffusion transformer so that the noisy action tokens can cross-attend to them~\cite{maniflow}. The point-cloud encoder (Utonia~\cite{utonia}) and text encoder (SigLIP~\cite{siglip}) are kept frozen; everything else is trained end-to-end with an MSE loss on the predicted $\epsilon$. During finetuning, LoRA adapters are trained for 2k steps; at deployment, the predicted actions are steered toward the nearest human trajectory. A short finetuning combined with action steering enables robust yet quick task adaptation.}
  \label{fig:architecture}
\end{figure*}

\subsection{Collecting and Processing Smart-Glass Data}
\label{sec:data}

\subsubsection{Data Collection}

\method{} uses two sources of egocentric human data, both collected by multiple operators wearing Aria Gen 2 smart glasses~\cite{ariagen2} at 10\,Hz. The first, $\mathcal{D}_{\mathrm{wild}}$, contains 109.6 hours of everyday interactions such as cooking, dishwashing, and cleaning, totaling 3.69M retained frames and 48.7k auto-generated instructions. The second, $\mathcal{D}_{\mathrm{task}}$, contains 5 task-specific demonstrations for each robot evaluation task. We use $\mathcal{D}_{\mathrm{wild}}$ for pretraining and $\mathcal{D}_{\mathrm{task}}$ for task-specific fine-tuning and action steering (i.e., \method{}).

Aria Gen 2 glasses provide synchronized RGB and stereo images, SLAM head pose, and per-hand tracking with 21 hand landmarks and a 6-DoF wrist pose. We additionally recover metric depth from the stereo images using S2M2~\cite{s2m2}. For each training snippet, we express the point cloud, hand trajectories, and head pose in the RGB camera frame of the last observed frame, giving a consistent camera-relative representation across demonstrations.

\subsubsection{Task Segmentation}
\label{sec:method:task_segmentation}
Raw smart-glass recordings are long and contain many interactions without explicit task boundaries or language annotations. Rather than asking a VLM to interpret an entire recording at once, we first split each video into short interaction snippets using hand--scene contact, then annotate each snippet with language and the manipulated object.

\paragraph{Contact-based segmentation}
We remove the demonstrator's arms and hands using SAM3~\cite{sam3}, unproject the remaining depth into a scene point cloud, mark a hand as active when its tracked landmarks come within $5$\,cm of the scene, and form snippets from intervals with a consistent active-contact state.

\paragraph{Language annotation.}
For each snippet, we sample up to 10 ordered frames and prompt Qwen3-VL-8B~\cite{qwenvl} to generate a short, verb-first instruction that describes the interaction without referring to the human embodiment, e.g., \emph{``placing bread in toaster''}.

\paragraph{Object grounding.}
We additionally condition the policy on a mask of the manipulated object, similar to~\cite{cap}. For each snippet, we prompt Gemini 2.5 Flash to place two keypoints on the contacted object, which initialize SAM3 to propagate the object mask through the snippet. Human arm and hand regions are removed from the resulting mask before it is used for training.

\subsubsection{Visual Domain Alignment}

Although task segmentation provides structured supervision, the visual observations still differ between human demonstrations and robot deployment: human arms and hands appear during training, while robot arms and hands appear at inference. We reduce this embodiment gap by removing the embodiment geometry from the observation in both settings.

During training, SAM3 provides the human arm and hand mask; at inference, the robot's joint configuration and kinematic model provide the corresponding robot geometry. In both cases, we remove the masked depth points before constructing the colored point cloud, so the policy observes the surrounding scene rather than either embodiment. We further restrict the point cloud to a camera-centered workspace defined by a $0.85\times1.5\times1.5$\,m ellipsoidal crop, discard points beyond $1.7$\,m in depth, and resample the remaining cloud to $5{,}000$ points. This yields a consistent observation representation between human training data and robot inference.

\subsection{Observation and Action Representation}
\label{sec:representation}

\paragraph{Coordinate frame}
We express each training snippet in the RGB camera frame at the anchor frame $t_a$, defined as the last observed frame before action prediction. The point cloud, hand history, future hand trajectories, and head targets are all transformed into this common frame using the Aria SLAM pose and camera extrinsics.

\paragraph{Action representation}
Each hand is represented by its 21 tracked landmarks plus three additional points encoding the wrist orientation, giving 24 points per hand. Four additional points represent the head pose, for a total of 52 points in $\mathbb{R}^3$. Each action chunk contains 15 future steps at stride 3, which at the 10\,Hz collection frequency corresponds to 4.5 seconds of motion.

\subsection{Pretraining and \method{} Policies from Human Videos}

Inspired by the success of diffusion transformers~\cite{dit} in robot learning~\cite{dit-policy, maniflow}, we train an $\epsilon$-predictor with a transformer backbone. Following~\cite{ddim}, we use $T=100$ diffusion timesteps during training and deterministic DDIM sampling ($\eta=0$) with $N=8$ steps at inference. Our architecture and the \method{} logic are illustrated in Fig.~\ref{fig:architecture}.

Similar to~\cite{maniflow}, observation embeddings condition the action prediction in two ways: (1) a context token, obtained by self-attention across all observation tokens, is added to each transformer block via AdaLN conditioning~\cite{dit, adaln}; and (2) the full observation-token sequence is fed to the transformer alongside the noisy action tokens, enabling cross-attention between the two. The resulting action tokens pass through a final feed-forward network (FFN) to predict the noise $\hat{\epsilon}$ at each denoising step. The loss is the mean-squared error between $\hat{\epsilon}$ and the ground-truth noise $\epsilon$.

We use four observation modalities: a cropped and sampled point cloud $P \in \mathbb{R}^{5000\times3}$, a binary manipulated-object mask $O \in \mathbb{R}^{255\times255}$, proprioception given by the 5 fingertip points of each hand $S \in \mathbb{R}^{2\times5\times3}$, and a task description $D$. The point cloud is encoded with Utonia~\cite{utonia} into $P_e \in \mathbb{R}^{512\times1024}$. The object mask is passed through a ResNet-18~\cite{resnet-18} trained from scratch to obtain a token $O_e \in \mathbb{R}^{1\times1024}$, proprioception through an FFN to obtain $S_e \in \mathbb{R}^{1\times1024}$, and the language annotation through a pretrained SigLIP~\cite{siglip} encoder to obtain $D_e \in \mathbb{R}^{1\times1024}$. These are concatenated into the observation-token sequence $Z = (P_e \,|\, O_e \,|\, S_e \,|\, D_e) \in \mathbb{R}^{515\times1024}$ and fed to the transformer. The Utonia and SigLIP encoders are kept frozen; everything else is trained end-to-end on $\mathcal{D}_{\mathrm{wild}}$.

\paragraph{Implementation details}
We apply per-head query/key normalization~\cite{qk-norm} for stable token mixing, and mean-pool the self-attended tokens to obtain the context latent used for AdaLN. Training hyperparameters are provided on our website.

\begin{figure*}[ht]
  \centering
  \includegraphics[width=\textwidth]{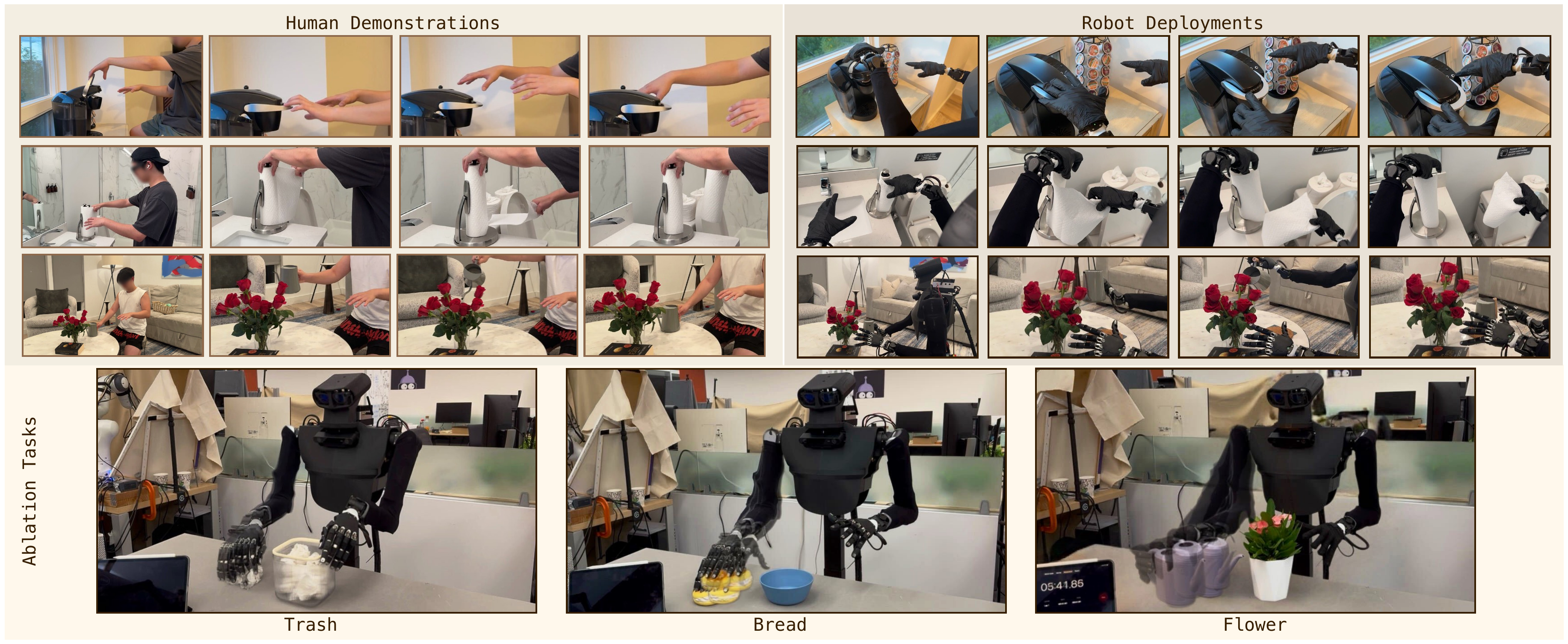}
  \caption{In-the-wild and ablation tasks. We deploy \method{} on 9 new tasks in new in-the-wild environments. For each task, we collect 5 new human demonstrations (top left) and deploy the policy on our robot YOR (top right). All ablation experiments are run on three tasks with spatial generalization (bottom). \method{} achieves an overall success rate of $71\%$ (7.1/10) across these 12 tasks; per-task success rates are given on our website.}
  \label{fig:tasks}
\end{figure*}

\begin{figure*}[ht]
  \centering
  \includegraphics[width=\textwidth]{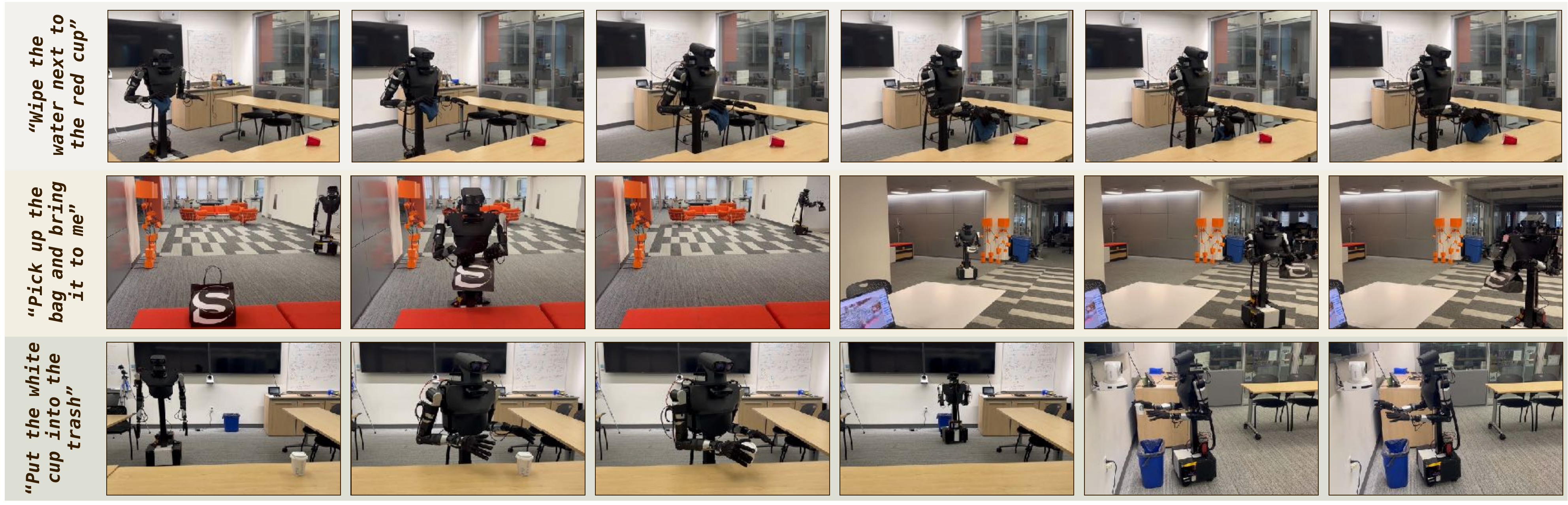}
  \caption{Full-stack deployments of different policies from a single language command. The left-most column shows the language annotations used by the system. For example, \textit{Put the white cup into the trash} is split into \texttt{navigate white cup}, \texttt{pick up white cup}, \texttt{navigate trash}, and \texttt{drop white cup}. Demonstrations for picking up and dropping the cup are collected and deployed in the same way as in \method{}.}
  \label{fig:full_stack_demos}
\end{figure*}

\subsubsection{Post-Training}
\label{sec:method:adaptation}

We explore different ways of enabling rapid task adaptation from only five new human demonstrations, and find that the best recipe is to combine fine-tuning with low-rank adapters~\cite{lora} and steering the noise distribution during reverse sampling with the provided demonstrations, similar to~\cite{demo-diffusion}. In both cases, the new demonstrations are processed as described in Section~\ref{sec:method:task_segmentation} and used as described below. Fig.~\ref{fig:budget-vs-steering} shows the impact of both on downstream robot tasks.

\paragraph{Low-Rank Adaptation (LoRA)}
Full fine-tuning is too flexible when only five demonstrations are available per task~\cite{tail}. We therefore freeze the pretrained weights and insert LoRA adapters~\cite{lora} of rank $8$ (scaling $16$) into the attention projections and selected linear layers. The action normalizer is loaded from pretraining so that post-training does not change the coordinate scale predicted by the denoiser. We fine-tune for only 2000 steps, totaling 2 minutes. To cover a broader distribution than the limited in-scene data provides, we augment the anchor frame, wrist pose, RGB, and text. We apply rigid transformations to the anchor frame $t_a$ and the proprioception points, with translations sampled uniformly within $\pm5$\,cm and $\pm3$\,cm and rotations within $\pm5^\circ$ and $\pm5^\circ$, respectively, for generalization to varying robot head and hand poses. We apply brightness, contrast, hue, saturation, and small noise jitter to the point-cloud RGB values. For text, we sample paraphrases that preserve the action while matching the short verb-object-preposition style of the pretraining interactions.

\paragraph{Action Steering}
While necessary, 2 minutes of fine-tuning is not sufficient to complete tasks robustly. Inspired by~\cite{repaint}, we therefore steer the action predictions during reverse sampling using the human actions from the demonstration whose initial frame is closest to the current observation. Specifically, at diffusion timestep $t$ and given a steering action $a^{\star}$, we first noise it to the same noise level using the forward diffusion process,
\begin{equation}
a^{\star}_t = \sqrt{\bar{\alpha}_{t}}\,a^{\star}_0 + \sqrt{1 - \bar{\alpha}_t}\epsilon,
\end{equation}
and then, given a steering weight $w$, shift the noisy actions as
\begin{equation}
x_t \leftarrow w\, a^{\star}_t + (1-w)\, x_t,
\end{equation}
before continuing with standard DDIM sampling.

To choose the closest human trajectory, we first re-express each demonstration scene in the manipulated-surface coordinate frame with the gravity axis pointing upward. We then compare the object layout in the current observation to those in the demonstrations and select the closest one. The object layout is extracted by segmenting each object on the table and computing its 3D centroid with respect to the camera. At each frame, we track the robot's progress through the selected demonstration by identifying the closest corresponding wrist and hand motion. This selection has a significant impact on success rate; we ablate it in Section~\ref{sec:experiments:trajectory_selection}.

\begin{figure*}[t]
\centering
\includegraphics[width=0.262\textwidth]{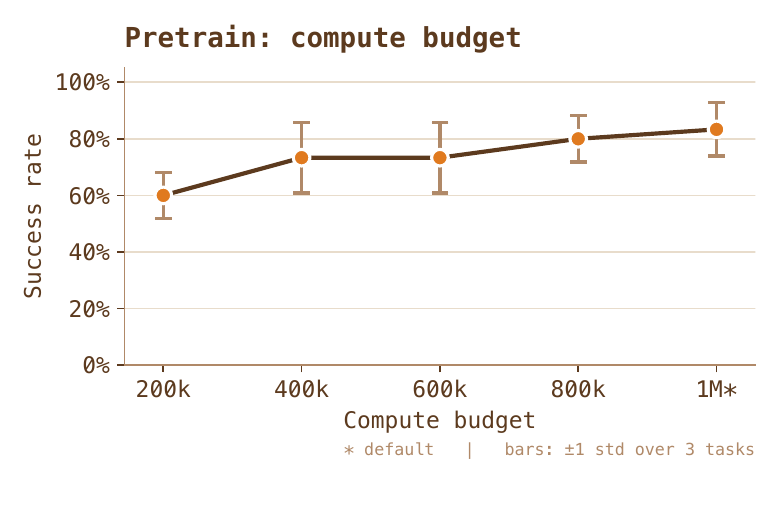}\hfill
\includegraphics[width=0.234\textwidth]{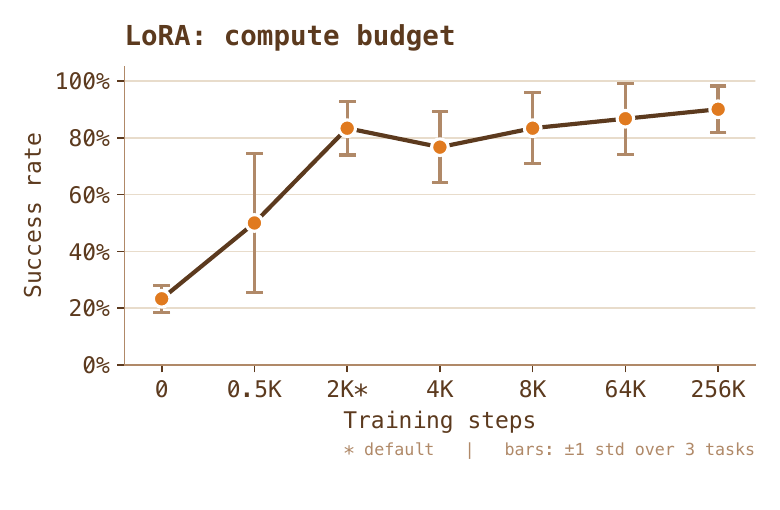}\hfill
\includegraphics[width=0.234\textwidth]{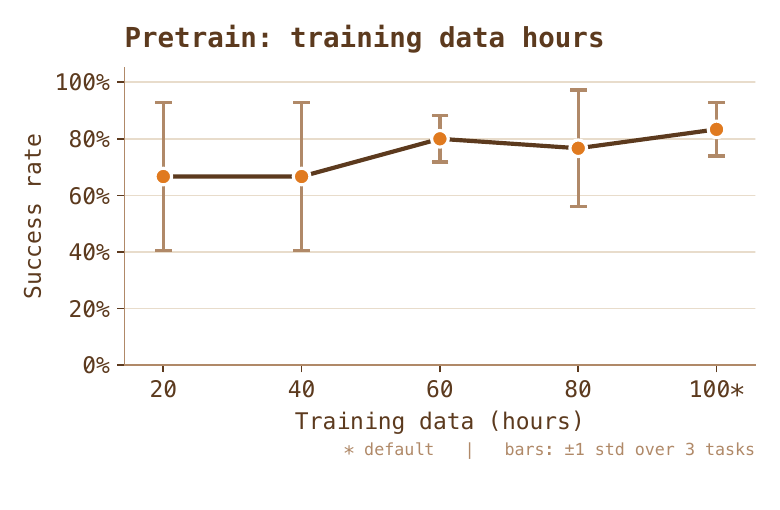}\hfill
\includegraphics[width=0.234\textwidth]{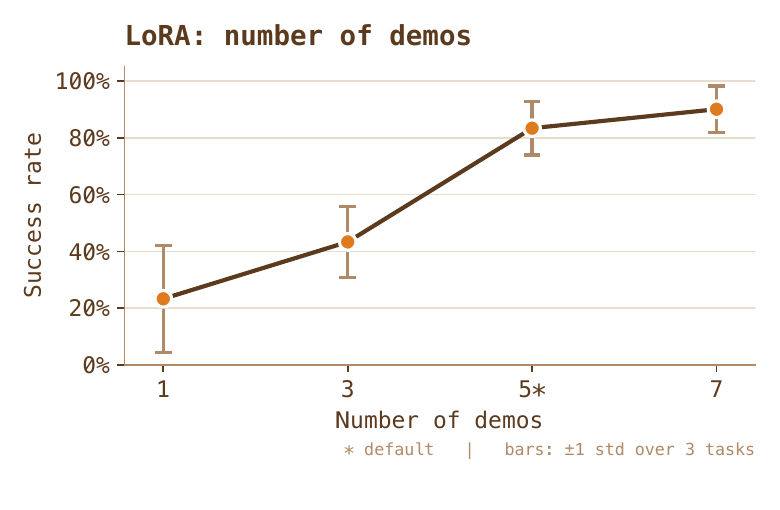}
\caption{Scaling ablations for (a)~pretraining compute budget, (b)~LoRA compute budget, (c)~pretraining data hours, and (d)~number of LoRA demonstrations. Points are the mean success rate over the three tasks; bars show $\pm1$ standard deviation. \method{} settings are marked with~*.}
\label{fig:scaling}
\end{figure*}

\subsection{Embodiment Transfer and Robot Deployment}

\subsubsection{Embodiment Transfer}

Our main robot embodiment is a YOR robot~\cite{yor} with Wuji hands~\cite{wuji} as end effectors. We attach the Aria glasses to the robot's head with a 3D-printed mount. During deployment, as with the human data, we remove the robot arm and hand from the observation point cloud using the robot's kinematic chain. We move the robot's wrists to the predicted wrist poses with Cartesian inverse kinematics and then apply kinematic retargeting~\cite{anyteleop} so the Wuji hands follow the predicted landmarks relative to the wrists. To minimize jitter, we apply real-time chunking~\cite{rtc} to temporally smooth the predictions.

\subsubsection{End-to-End Deployment}

To showcase the generalization of \method{}, we build an end-to-end platform that lets the robot complete tasks in new environments from a language command. We use Gemini ER 1.6~\cite{gemini} to decompose a long-horizon command into $\texttt{navigate <object>}$ and $\texttt{manipulate <task>}$ sub-tasks. For $\texttt{navigate}$, we follow~\cite{ok-robot}: we extract a voxelized SLAM map of the environment in which each voxel is associated with a SAM3~\cite{sam3} latent, find the nearest neighbor of the target object's SAM3 latent in the map, and navigate the robot to that location. Once there, the $\texttt{manipulate}$ sub-task runs the multi-task manipulation policy described above. Fig.~\ref{fig:full_stack_demos} shows a few end-to-end tasks with this setup; robot rollouts are on our website.
\section{Experiments}

% \method{} is an end-to-end system for learning in-the-wild robot policies that can be deployed in homes and adapted to new tasks in under 5 minutes of processing time.
In this section, we run a series of robot experiments to answer the following questions:
\begin{enumerate}
    \item What is the impact of fine-tuning and action steering on quick task adaptation?
    \item What is the impact of the training budget (i.e., training steps and data) on downstream robot policies?
    \item How well does \method{} perform when ablated against different architectural components?
    \item How much does the trajectory-selection strategy used during action steering impact downstream task performance?
    \item How well does \method{} generalize to new objects?
\end{enumerate}

\subsection{Task Descriptions}

We deploy \method{} on 12 tasks in total, illustrated in Fig.~\ref{fig:tasks}: 9 in an Airbnb rented for this purpose and 3 in our lab. For each task, we collect 5 entirely new human demonstrations and deploy the policy immediately afterwards. All ablation experiments are run on the 3 lab tasks. In \textit{Trash} and \textit{Bread}, the robot is asked to pick up the corresponding object and drop it into a bowl or a can, respectively. In \textit{Flower}, the robot is asked to pick up the watering can and water the flower. During evaluation, we move the trash, bread, and watering can within a $20\times20$\,cm region. \method{} achieves an overall success rate of $71\%$ (7.1/10) across these 12 tasks. All robot videos are available on our website.

\subsection{What is the impact of fine-tuning and action steering on quick task adaptation?}
\label{sec:experiments:steertuning}

\begin{figure}[h]
    \centering
    \includegraphics[width=\linewidth]{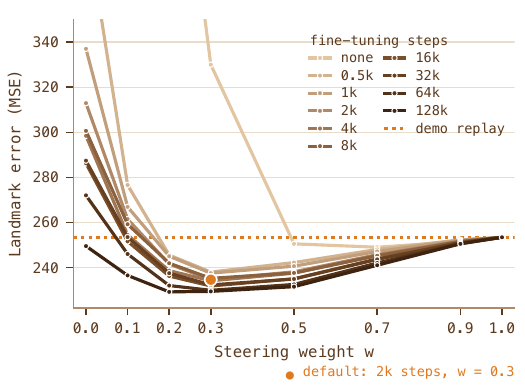}
    \caption{Mean landmark error as a function of the steering weight $w$ for different fine-tuning budgets, averaged over our 12 tasks. Error is the MSE between the predicted and ground-truth action landmarks on a held-out demonstration; the dotted line is the error of replaying the demonstration directly. Values above 350 (no fine-tuning at $w \le 0.2$, 0.5k steps at $w=0$) are cut off for readability.}
    \label{fig:budget-vs-steering}
\end{figure}

\method{} uses fine-tuning and diffusion steering together in a novel way. Fig.~\ref{fig:budget-vs-steering} shows how the mean action error changes with longer fine-tuning and a higher steering weight $w$ on a held-out demonstration. The dotted line is the mean error of the trajectory selected to steer the policy on that held-out configuration. As we steer more strongly toward that demonstration, the error converges to the dotted line (at $w=1$, DDIM sampling reduces to replaying the demonstration). Darker lines correspond to policies fine-tuned for more steps, from no fine-tuning to 128k steps. While several configurations bring the error below the dotted line, fine-tuning for only 2k steps (about 2 minutes) and steering with $w=0.3$ strikes the best balance: it reaches a minimum error very close to that of the policy fine-tuned for 128k steps (roughly a day of training) at a small fraction of the cost.

\begin{figure*}[h]
    \centering
    \includegraphics[width=\linewidth]{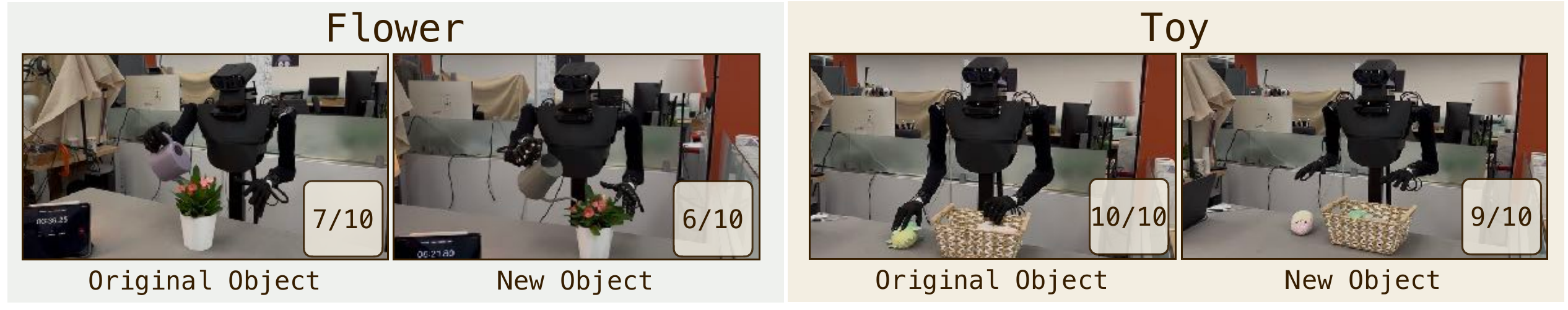}
    \caption{Object generalization. For each task, the left image shows the original object and the right image the new object used at deployment, with the corresponding success rate.}
    \label{fig:object_generalization}
\end{figure*}

\begin{table}[h]
  \centering
  \caption{Success rates of the full \method{} pipeline compared to fine-tuning only and steering only. Ours in bold.}
  \label{tab:finetuning_steering}
  \footnotesize\setlength{\tabcolsep}{2.5pt}
  \begin{tabular}{lcccc}
    \toprule
    \textbf{Pipeline} & \textbf{Trash} & \textbf{Flower} & \textbf{Bread} & \textbf{Avg.} \\
    \midrule
    \textbf{\method{}} & \sr{90} & \sr{70} & \sr{90} & \sr{83} \\
    Steering only & \sr{20} & \sr{20} & \sr{30} & \sr{23} \\
    Short fine-tune (2k) & \sr{0} & \sr{0} & \sr{0} & \sr{0} \\
    Full fine-tune (150k) & \sr{70} & \sr{80} & \sr{80} & \sr{77} \\
    \bottomrule
  \end{tabular}
\end{table}

Beyond landmark errors, we run robot experiments on the 3 lab tasks and compare \method{} against three baselines: fine-tuning only, for 2k and for 150k steps, and steering only, without any fine-tuning (Table~\ref{tab:finetuning_steering}). \method{} outperforms all baselines. The closest, fine-tuning for 150k steps, reaches comparable success rates but requires a full day of training, whereas \method{} adapts in minutes. The same 2k-step fine-tuning that \method{} uses fails completely on its own, and steering alone succeeds in only $23\%$ of trials. Together, these results show that combining the two is what enables few-shot task adaptation that is both robust and fast.

\subsection{What is the impact of the training budget (i.e., training steps and data) on downstream robot policies?}

\method{} brings together many data sources and methodologies for learning robot policies. To quantify the impact of data and compute, we vary the compute budget for both pretraining and LoRA fine-tuning, as well as the amount of data used in each stage, and report success rates with standard-deviation bars in Fig.~\ref{fig:scaling}. Success rate increases with compute budget. For pretraining, we use the maximum budget to reach the highest performance; for fine-tuning, we instead choose the cheapest checkpoint that still achieves a high success rate, keeping task adaptation as fast as possible.

\subsection{How well does \method{} perform when ablated against different architectural components?}

\method{} uses Utonia~\cite{utonia} as the point-cloud encoder and a Diffusion Transformer (DiT)~\cite{dit} as the action decoder. We ablate both by replacing the encoder with PointNeXt~\cite{pointnext} or PTv3~\cite{ptv3}, and the decoder with a UNet~\cite{unet}. Pretraining, post-training, and evaluation are identical for all variants. As shown in Table~\ref{tab:architecture}, Utonia and DiT achieve the highest average success rates; although other variants outperform \method{} on individual tasks, the averages support our design choices.

\begin{table}[h]
  \centering
  \caption{Success rates for encoder and decoder architecture ablations. Ours in bold.}
  \label{tab:architecture}
  \begin{tabular}{llcccc}
    \toprule
    & & \textbf{Trash} & \textbf{Flower} & \textbf{Bread} & \textbf{Avg.} \\
    \midrule
    \multirow{3}{*}{Encoder}
      & PointNeXt              & \sr{50}  & \sr{60} & \sr{70} & \sr{60} \\
      & PTv3                   & \sr{100} & \sr{40} & \sr{90} & \sr{77} \\
      & \textbf{Utonia (Ours)} & \sr{90}  & \sr{70} & \sr{90} & \sr{83} \\
    \midrule
    \multirow{2}{*}{Decoder}
      & UNet                   & \sr{80}  & \sr{80} & \sr{60} & \sr{73} \\
      & \textbf{DiT (Ours)}    & \sr{90}  & \sr{70} & \sr{90} & \sr{83} \\
    \bottomrule
  \end{tabular}
\end{table}

\subsection{How much does the trajectory-selection strategy used during action steering impact downstream task performance?}
\label{sec:experiments:trajectory_selection}

\method{} selects the nearest neighbor among the newly collected human demonstrations by comparing the object layout at deployment time to that of each demonstration, as described in Sec.~\ref{sec:method:adaptation}. We compare this against two alternatives for choosing the steering trajectory (Table~\ref{tab:trajectory_selection}):
\begin{itemize}
    \item \textbf{VINN}~\cite{vinn} uses the mean-pooled context latent of the initial observation to find the nearest-neighbor demonstrations, then takes a weighted average of their actions, with weights given by the normalized L2 distances between the context latents of the deployment and demonstration observations.
    \item \textbf{Wrist pose} uses the initial wrist pose to find the nearest-neighbor demonstration.
\end{itemize}

\begin{table}[h]
  \centering
  \caption{Success rates for trajectory-selection strategies at inference. Ours in bold.}
  \label{tab:trajectory_selection}
  \begin{tabular}{lcccc}
    \toprule
    \textbf{Trajectory selection} & \textbf{Trash} & \textbf{Flower} & \textbf{Bread} & \textbf{Avg.} \\
    \midrule
    VINN~\cite{vinn}             & \sr{70} & \sr{20} & \sr{70} & \sr{53} \\
    Wrist pose                   & \sr{60} & \sr{70} & \sr{40} & \sr{57} \\
    \textbf{Object-based (Ours)} & \sr{90} & \sr{70} & \sr{90} & \sr{83} \\
    \bottomrule
  \end{tabular}
\end{table}

Once a trajectory is selected, steering is performed identically for all methods. As shown in Table~\ref{tab:trajectory_selection}, the object-based selection of \method{} outperforms both alternatives.

\subsection{How well does \method{} generalize to new objects?}

Finally, we evaluate \method{} when the object at deployment differs from the one in the demonstrations. Results and the objects used are shown in Fig.~\ref{fig:object_generalization}. The \textit{Toy} task, where the goal is to pick up a toy and drop it into a box, was added specifically for this evaluation. \method{} remains robust to visual changes in the object as long as the required action does not differ significantly from the demonstrated one. We attribute this to large-scale pretraining and to the augmentations applied during fine-tuning.
\section{Discussion and Limitations}

We presented \method{}, a system for few-shot task adaptation from in-the-wild human videos in under 5 minutes. Pretraining provides robustness to visual perturbations, while the combination of fine-tuning and action steering enables rapid learning of new tasks in new environments. We note two major limitations. (1) Human video data contains no contact information; all actions and hand poses are annotated purely visually. As a result, the learned low-level actions remain relatively primitive, which limits the dexterity of the system. Incorporating contact information or a form of RL-based adaptation should extend the range of tasks \method{} can learn. (2) \method{} relies on kinematic retargeting being sufficient to complete the tasks; even when the human-policy predictions are accurate, the robot can fail due to retargeting errors.
\section*{Acknowledgements}
We thank Jeff Cui, Zifan Zhao, Dandan Shan, Bowen Tan, Andre Wang, Tianxing Zhou, Liangguang Zhou, Zicheng Teng, Lucas Liu, Sissi Hu, and Charles Ji for valuable discussions and feedback, and Mais Dwal, Kikki Liu, Jingyu Li, Melody Zuo, Kevin Tang, Ayaan Mustafa, Neytiri Cheng, Wilfendjy Mathurin, Maggie Wang, Cindy Ma, John Mwangi, Chloe Wang, Tasnim Raisa, Amelia Tang, and Bruce Zhang for their help with data collection.

This work was supported by grants from LG, Qualcomm, Honda, Microsoft, NSF award 2339096, ONR award N00014-22-1-2773 and Institute of Information \& Communications Technology Planning \& Evaluation (IITP) with a grant funded by the Ministry of Science and ICT (MSIT) of the Republic of Korea in connection with the Global AI Frontier Lab International Collaborative Research Agreement. (No. RS-2024-00469482 \& RS-2024-00509279). Lerrel Pinto is supported by the Sloan, Packard, and CIFAR Fellowships. 

Artificial intelligence (AI) tools were used to improve the grammar and clarity of this paper. AI was not used to generate any of the text or references.

% References
% {\footnotesize
% \bibliographystyle{IEEEtran}
% \bibliography{ref}
% }
% {\scriptsize
% \bibliographystyle{IEEEtran}
% \bibliography{ref}
% }

{\renewcommand{\footnotesize}{\scriptsize}%
\bibliographystyle{IEEEtran}
\bibliography{ref}
}

\end{document}